\documentclass[runningheads]{llncs}
\usepackage{graphicx}
\usepackage{amsmath,amssymb} 
\usepackage{color}
\usepackage{booktabs}
\usepackage{multirow}
\usepackage{hyperref}

\begin{document}

\newcommand{\point}{
    \raise0.7ex\hbox{.}
    }


\pagestyle{headings}

\mainmatter

\title{Visual Speech Recognition Using PCA Networks and LSTMs in a Tandem GMM-HMM System} 

\titlerunning{VSR Using PCA Networks and LSTMs in a Tandem GMM-HMM System} 

\authorrunning{M.~Zimmermann, M.~Mehdipour Ghazi, H.~K.~Ekenel, J.-P.~Thiran} 

\author{Marina Zimmermann$^1$, Mostafa Mehdipour Ghazi$^2$, Haz{\i}m Kemal Ekenel$^3$, Jean-Philippe Thiran$^1$} 
\institute{
$^1$ Signal Processing Laboratory (LTS5), Ecole Polytechnique F\'{e}d\'{e}rale de Lausanne (EPFL), Lausanne, Switzerland\\
$^2$ Faculty of Engineering and Natural Sciences, Sabanci University, Istanbul, Turkey\\
$^3$ Department of Computer Engineering, Istanbul Technical University (ITU), Istanbul, Turkey\\
marina.zimmermann@epfl.ch, mehdipour@sabanciuniv.edu, ekenel@itu.edu.tr, jean-philippe.thiran@epfl.ch} 

\maketitle

\begin{abstract}

Automatic visual speech recognition is an interesting problem in pattern recognition especially when audio data is noisy or not readily available. It is also a very challenging task mainly because of the lower amount of information in the visual articulations compared to the audible utterance. In this work, principle component analysis is applied to the image patches --- extracted from the video data --- to learn the weights of a two-stage convolutional network. Block histograms are then extracted as the unsupervised learning features. These features are employed to learn a recurrent neural network with a set of long short-term memory cells to obtain spatiotemporal features. Finally, the obtained features are used in a tandem GMM-HMM system for speech recognition. Our results show that the proposed method has  outperformed the baseline techniques applied to the OuluVS2 audiovisual database for phrase recognition with the frontal view cross-validation and testing sentence correctness reaching 79\% and 73\%, respectively, as compared to the baseline of 74\% on cross-validation.

\end{abstract}


The final publication is available at Springer via \url{http://dx.doi.org/10.1007/978-3-319-54427-4_20}.

\section{Introduction}

Visual speech recognition has seen increasing attention in recent decades. The research interest in this topic arises from several factors; first, visual speech recognition can be used in automatic audio-visual speech recognition together with the audio data \cite{Potamianos2004}. It has been shown that highly noisy audio data can thus be supported and higher recognition rates can be achieved in such scenarios \cite{Potamianos2004}. In these cases, the supporting visual information helps similarly to its contribution in human-human interaction, where a listener's concentration on the lip movement increases in noisy environments. Secondly, visual speech recognition (VSR) is an interesting topic with varied applications. To name just a few, cybersecurity (pronounced passwords) \cite{Hassanat2014}, sign language (accompanying mouthings) \cite{Schmidt2013}, speech production \cite{Badin_2002} and in general human machine interfaces have an interest in VSR.

While audio-based speech recognition has improved significantly over the past decades and is nowadays applicable in many real-life scenarios, visual speech recognition still mostly focuses on speech produced in controlled lab conditions. However, there is a lot of interest to address, for example, the problem of head pose, which is a large hindrance in the application to real-world scenarios. Various works have already addressed this problem by taking different view angles into account \cite{Lucey2007,Estellers_2012}. To this end, several databases have been recorded simultaneously with cameras at different angles \cite{Lee2004,Harte_2015}. The recently published OuluVS2 database \cite{Anina_2015} aims at being a comparative dataset to allow a comprehensive comparison of approaches on multi-view data where cameras at five different angles record a subject simultaneously.

Efforts to bring visual speech recognition up to date with novel techniques used in both audio speech recognition and computer vision direct researchers to utilize deep learning techniques. Deep neural networks (DNNs) are widely employed in audio-based automatic speech recognition resulting in the current baseline accuracies \cite{graves2014}. DNNs have also become the standard techniques in computer vision to set baselines in recognition or analysis tasks \cite{Donahue_2015,Chan_2015}. However, one big problem in applying these networks to visual speech data is the fact that visual speech databases are not comparable to audio databases in terms of their sizes and number of speakers, meaning that insufficient amounts of training data are available. This is an important drawback since having a large amount of data is a necessity for training deep learning frameworks for complex acoustic models and complete recognition chains used for continuous speech. Although a few larger audio-visual databases such as TCD-TIMIT and OuluVS2 \cite{Harte_2015,Anina_2015} have been published recently, the problem still remains highly challenging.

In this paper, we propose a visual speech recognition approach based on a two-stage PCA-based convolutional network \cite{Chan_2015} followed by a layer of long short-term memories (LSTMs) to extract a set of unsupervised spatiotemporal visual features. These features are then used in a tandem GMM-HMM system for speech recognition. Our contribution is two fold, with a major focus on feature extraction. First, we use principal component analysis in a multi-stage convolutional network to extract the optimal unsupervised learning lip representations. Secondly, we apply recurrent neural networks (RNNs) with LSTM cells to lip representations to extract spatiotemporal features. This approach does not only find the time-series dependencies within the video frame sequences, but also decreases the lips feature set dimension for further processing with the GMM-HMM scheme. Using this system, we were able to improve the baseline cross-validation results for phrase recognition for this workshop from a frontal and $30^\circ$ side view with a large margin of roughly 5\%, reaching 79\% of all sentences being recognized correctly for each of these views. Combining these two views leads to an even higher recognition rate of 83\% of all sentences.

The rest of this paper is organized as follows. Section 2 briefly reviews the related work and state-of-the-art approaches. Section 3 explains the details of the proposed method for visual speech recognition based on a PCA network, LSTMs, and the GMM-HMM system. Section 4 describes the utilized dataset, experiments, and obtained results. Finally, Section 5 concludes the paper with a summary and discussions.

\section{Related Work}

Visual speech recognition requires a series of steps to process the video and extract relevant features. First of all, a region of interest (ROI) around the mouth, which contains the largest amount of information about the utterance, has to be extracted \cite{Potamianos2004}. This can be done by hand or with the help of a face tracker. The latter is more common nowadays even though manual corrections are still sometimes applied. The ROI is later used to extract the features.

In general three types of features are used: texture-based features, shape-based features, or a combination of both \cite{Potamianos2004,Bowden_2013}. Texture-based features exploit the pixel values in a ROI --- usually closely around the mouth or including the jaws \cite{Potamianos2004}. Typically, this is done by applying a transformation such as the discrete cosine transform (DCT) and/or a dimensionality reduction technique such as the linear discriminant analysis (LDA) to the ROI, possibly in combination with a principle component analysis (PCA) or a maximum-likelihood linear transform (MLLT) \cite{Potamianos2004}. A common feature post-processing technique involves a chain of LDAs and MLLTs on concatenated frames, the so-called HiLDA \cite{Potamianos_2003}.

Shape-based features, on the other hand, try to extract information about the shape of the mouth. This can be done for example with the help of snakes, taking into account the outer contours of the mouth, or by computing the geometrical distances between certain points of interest around the mouth \cite{Potamianos2004}. In recent works these feature points are generally extracted with the help of a face or mouth tracker. Some researchers also directly use these points or shapes and extract information by applying a PCA to them. This technique is, for example, the case for the use of active appearance models (AAMs) \cite{Bowden_2013,Biswas_2015}.

The next step in the recognition system is the classification of the utterance, traditionally performed through a system composed of Hidden Markov Models (HMMs) with Gaussian Mixture Models (GMMs). The GMMs model the acoustics, i.e.\ the phonemes, or visemes in visual speech, while the states of an HMM model the time evolution within a phoneme and the overall evolution within and between words \cite{Potamianos2004}.

Even though there are still many studies working on texture-based features such as DCT, DCT-HiLDA, or scattering \cite{Mroueh_2015} and, similarly, many researchers still work with GMM-HMM recognition systems, recently, more focus is being put on deep learning techniques. These networks are widely spread both in audio speech recognition and visual recognition tasks to extract features, construct acoustic models, or replace the complete recognition chain.

In recent literature, deep network based approaches have consistently shown superior performances over traditional methods. Deep Boltzmann machines have been used as stacked autoencoders for feature extraction \cite{Ngiam2011} or post-processing of local binary patterns from three orthogonal planes (LBP-TOP) \cite{Sui_2015}. These features are then classified using Support Vector Machines (SVMs) \cite{Ngiam2011}, where all utterance lengths have to be normalized, or using a tandem system \cite{Hermansky}, where the features are passed into a GMM-HMM recognizer \cite{Huang_2013,Noda_2014,Sui_2015}. Similarly, feature extraction has been performed by convolutional neural networks (CNNs) \cite{Noda_2014,Koller_2015} and deep belief networks (DBNs) \cite{Huang_2013}. The outputs of these networks can be used as an acoustic model in the so-called hybrid approach, where the posterior probability outputs are passed directly to the HMM \cite{Bourlard_1994}. Finally, the recognition system itself can be replaced by DNNs, either in the form of bilinear \cite{Mroueh_2015} or recurrent neural networks \cite{Wand_2016}. In the former case, DNNs are used to classify texture-based features while in the latter case the whole processing chain is replaced by a LSTM network.

Comparing these approaches to our proposed method, one speciality of the PCA network is the effectiveness with which it extracts information from the given frames without needing any prior knowledge --- it is unsupervised. The two-stage projection onto the leading principle components allows to capture the main variations within the image patches, while also extracting higher level features through the concatenation of these in two stages. The subsequent binarization and extraction of histograms leads to an indexing and pooling of these, a non-linear step. The following LSTM network then reduces the feature size dramatically by taking into account the temporal information between different frames and the classes assigned to these. As a result, the posterior probabilities serve as good spatiotemporal descriptors and can be utilized in a tandem system as features for a GMM-HMM recognizer.

\begin{figure}
\begin{center}
\includegraphics[width=0.84\columnwidth]{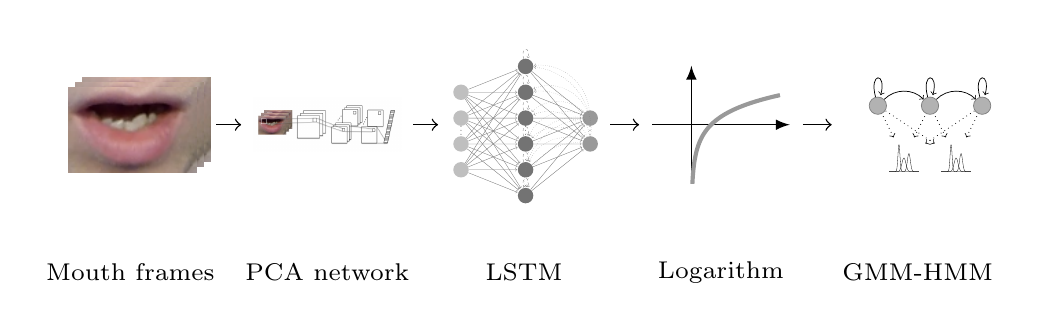}
\caption{{\label{fig:flowchart}
The proposed method for visual speech recognition from the mouth video frames.%
}}
\end{center}
\end{figure}

\section{The Proposed Method}

In this section, novel feature extraction methods are explored for visual speech recognition. More specifically, a two-stage PCA-based convolutional network \cite{Chan_2015} followed by a layer of LSTMs \cite{Hochreiter_1997} extracts features from the cropped mouth images. The obtained spatiotemporal features are then processed in a tandem system with a GMM-HMM basis for speech recognition.

\begin{figure}
\begin{center}
\includegraphics[trim=3.7cm 4.2cm 2cm 3.5cm, clip,width=0.84\columnwidth]{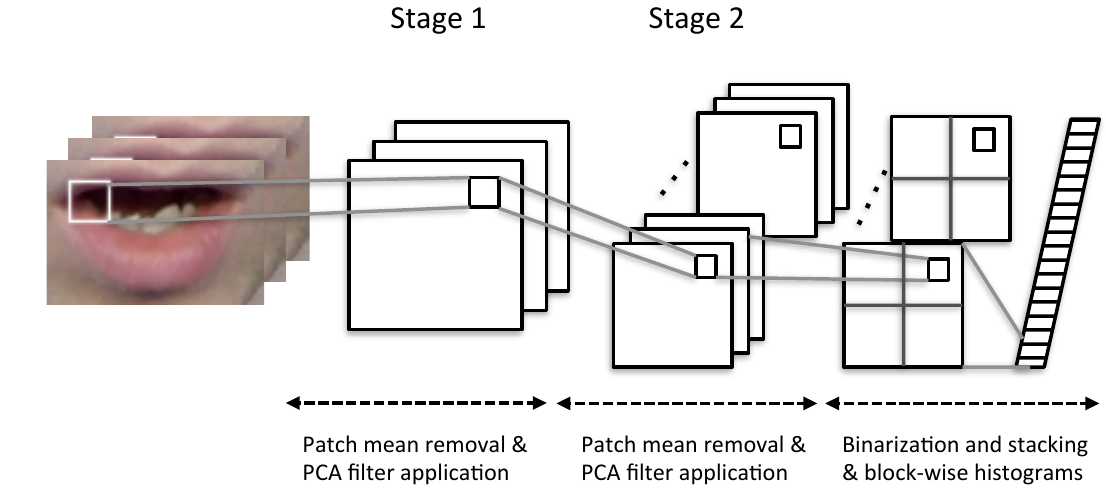}
\caption{{\label{fig:pcanet}
The PCA network used in the first stage of the proposed method.%
}}
\end{center}
\end{figure}

Feature extraction is performed in a sequential fashion as shown in Figure~\ref{fig:flowchart}. First, a two-stage PCA network is applied to each video frame (see Figure~\ref{fig:pcanet}). The first layer network weights are learned by applying PCA to concatenated square patches --- which are extracted from the mouth video frames and then vectorized. We use eight principal components as the networks' first layer filter bank and convolve these with the input images. In a cascaded scheme, a similar procedure is applied to the filtered patches to obtain the second layer filter bank. After convolution, the output maps are binarized with a Heaviside step function and every eight binary images are stacked together to compose an 8-bit image --- similar to the first layer outputs. Finally, block histograms --- with 256 bins --- are extracted from the obtained maps and concatenated, resulting in a long feature vector for each video frame. In this work, we extract 16 block histograms which result in feature vectors of length $16 \times 256 \times 8 = 32,768$.

Secondly, an LSTM network is connected to the outputs of the PCA network to extract more abstract representations while taking the time-series dependencies between the video frames into account. This type of RNN is composed of memory cells to store the past values or ignore the dependencies when needed. Therefore, each cell has an input, an output, and a forget gate that can be activated at different levels. This architecture results in three cases: accepting the new input value, forgetting the existing value, or outputting a value at the given level \cite{Graves_2013}. Since we label each video frame in the phrase recognition subset based on the audio phonemes, there are 28 output nodes in our LSTM network.

Last but not least, the posterior probabilities received from the LSTMs are passed as spatiotemporal features concatenated with their delta and acceleration components into a GMM-HMM based speech recognition system, the so-called tandem approach. This system is implemented using the Hidden Markov Model Toolkit (HTK) \cite{Young2002}. However, since the outputs of the LSTM network show small variations, we first take the logarithm of these features to make them more discriminative. Our tandem system contains GMMs with 15 Gaussian mixtures per observation and 4 states per word.

\section{Performance Analysis}

In this section, we review details of the utilized dataset, evaluation metrics, and the conducted experiments. We present the validation and test results and discuss them in detail.

\subsection{The Dataset}

We use the phrase recognition subset of the OuluVS2 database \cite{Anina_2015} in our experiments. This dataset contains video clips of 52 subjects from five different views: frontal and four side views at $30^\circ$, $45^\circ$, $60^\circ$, and $90^\circ$ (the profile). During each recording session, the subjects were asked to utter 10 daily short English phrases shown on a computer monitor. Each phrase was repeated three times resulting in 30 video recordings (utterances) per subject per view. The recording was performed in an ordinary office environment with varying lighting conditions and background noises producing a more real-world audio-visual dataset. Each of these videos was recorded with a resolution of $1920 \times 1080$, at 30~fps, and with an audio bit rate of 128~kbps. The challenge organizers also provided aligned and cropped mouth videos along with the original videos and fixed the training and test subsets: 40 out of 52 subjects are assigned for training and the rest are used for testing.

\begin{figure}
\begin{center}
\includegraphics[width=0.7\columnwidth]{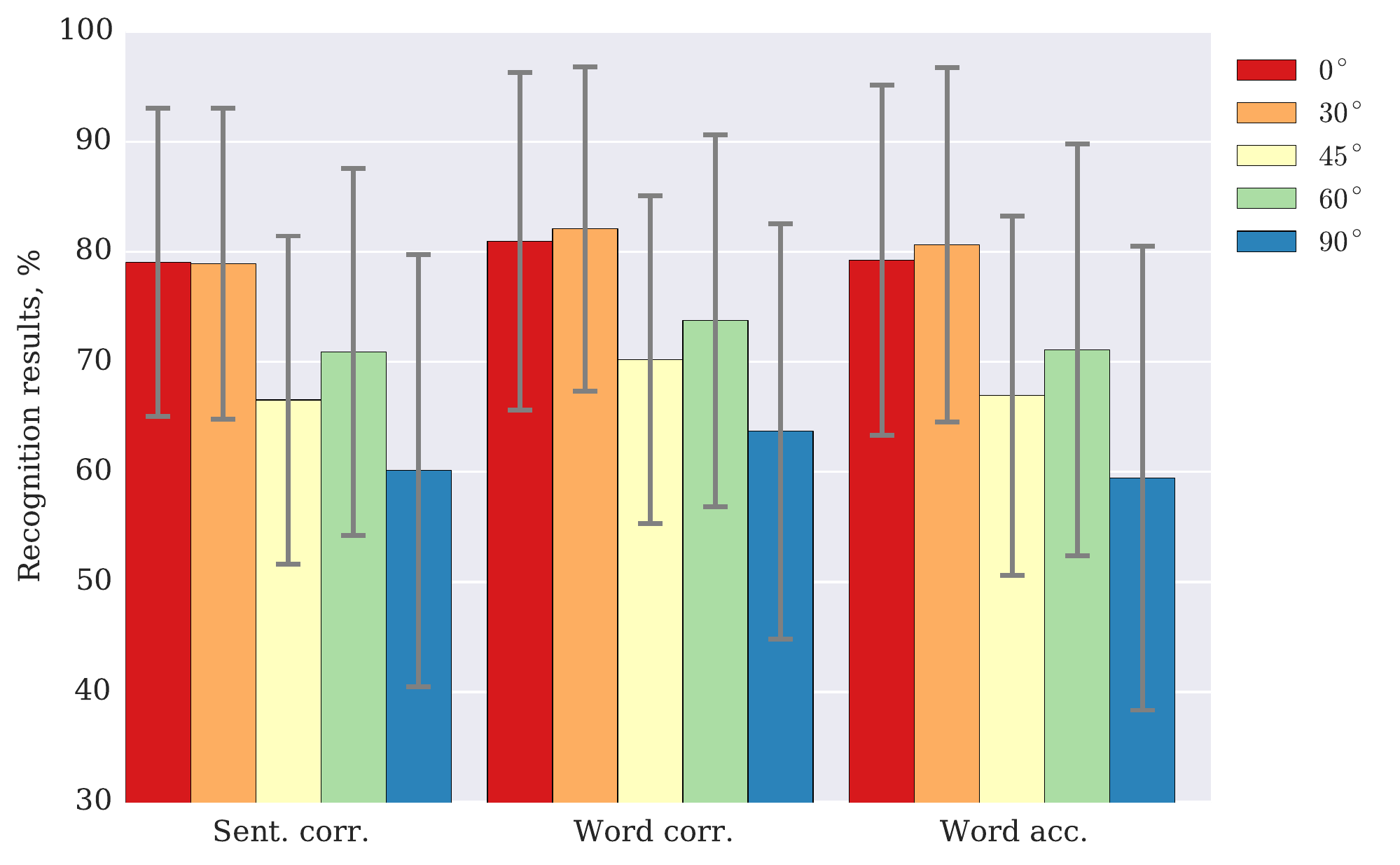}
\caption{{\label{fig:exp4}
Mean phrase recognition results on the multi-view dataset of OuluVS2 using our proposed method with the cross-validation technique and the standard deviation across subjects.%
}}
\end{center}
\end{figure}

\begin{figure}
\begin{center}
\includegraphics[width=0.7\columnwidth]{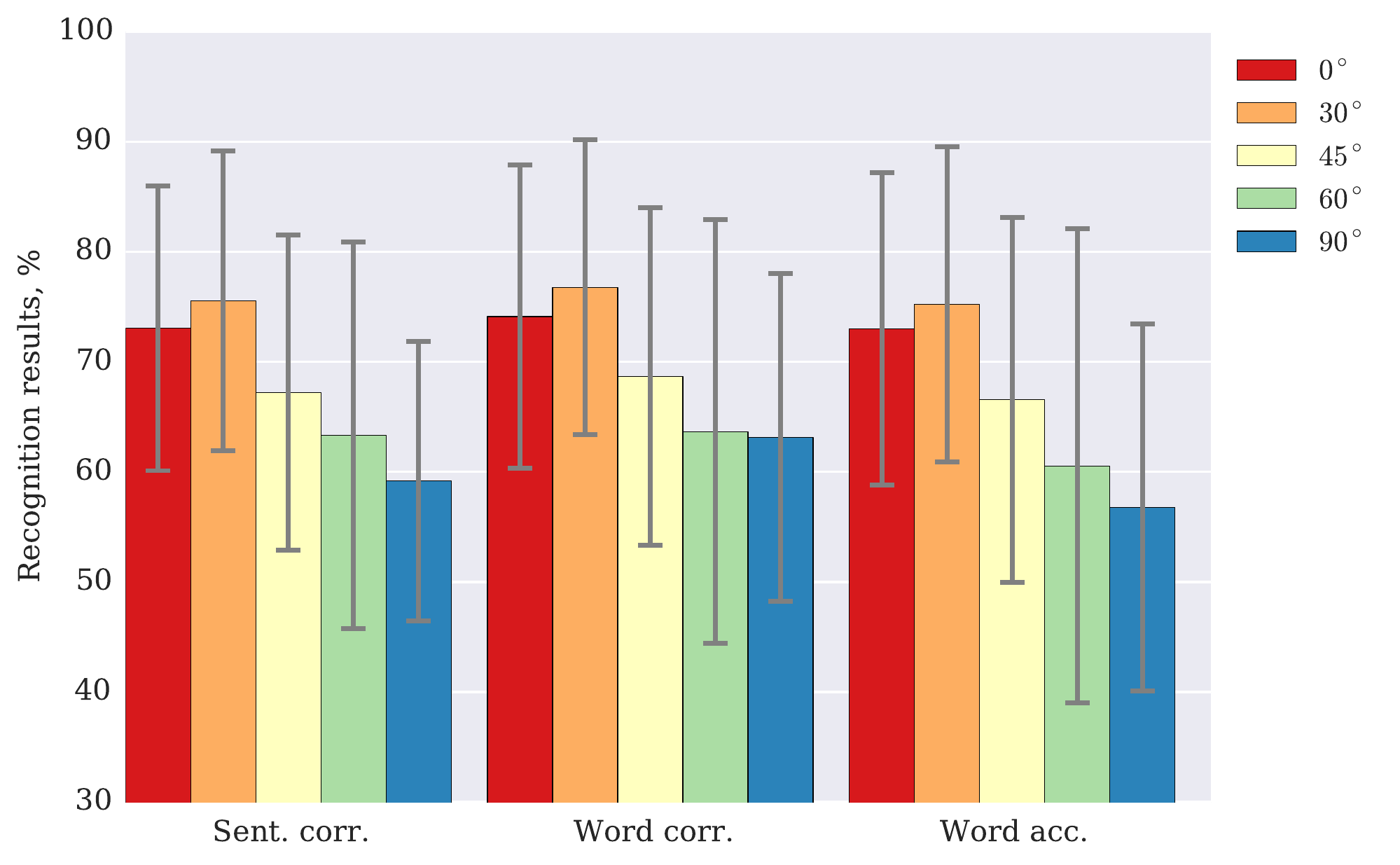}
\caption{{\label{fig:exp4_test}
Mean phrase recognition results on the multi-view dataset of OuluVS2 using our proposed method on the given test set and the standard deviation across subjects.%
}}
\end{center}
\end{figure}

\subsection{Experimental Results}

In our experiments, we use the provided cropped mouth videos by first extracting and converting all video frames to grayscale images of size $60 \times 90$ pixels. PCA is applied to all image patches of size $7 \times 7$ pixels to learn eight filter banks in a two-stage cascaded PCA network. We add a max-pooling layer to the output of this network to obtain a more abstract representation before histogram pooling. Finally, 16 block histograms are extracted and concatenated to obtain a 32,768-dimensional feature vector for each frame.

For spatiotemporal recognition using the LSTM network, we need to obtain frame-based labels using phoneme level transcription. For this purpose, the audio data is first aligned to the sentence transcriptions using a standard GMM-HMM system with MFCCs trained on the training subset. These transcriptions are then used as labels for the obtained feature set from the PCA network. We train a one-layer LSTM network with a Sigmoid activation function in the gates and cells. The learning rate, weight decay penalty, and momentum value are set to 0.5, 0.001, and 0.8, respectively. Moreover, we use a random batch size and train the network until 10,000 iterations.

Three metrics are used to present the results: the accuracy and correctness at the word level, and the percentage of correct sentences. The word accuracy and correctness are defined as follows

\begin{equation}
    \mathrm{Accuracy} = \frac{H - I}{N} \cdot 100\%
\end{equation}
\begin{equation}
    \mathrm{Correctness} = \frac{H}{N} \cdot 100\%
\end{equation}
where $H$, $I$, and $N$ are the number of correct words, number of erroneous words (insertion error), and the total number of words, respectively. The number of correct words is equal to the number of all words minus the total number of ignored words (deletion error) and the number of wrongly recognized words (substitution error), i.e.\ $H = N - D - S$.

To adjust our system parameters, we use a leave-one-out cross-validation scheme on the given training set. Later on, we apply the system in a leave-one out cross-validation scheme on the whole data, similar to the baseline\footnote{The baseline results can be found at \url{http://ouluvs2.cse.oulu.fi/preliminary.html}}, and, trained on the training set, to the test set for the final recognition at the word or phrase levels. Figures~\ref{fig:exp4} and \ref{fig:exp4_test} show our cross-validation and test results on the OuluVS2 dataset for phrase recognition.

\setlength{\tabcolsep}{1.5pt}
\begin{table}
\caption{
\label{tab:exp4_test} Phrase recognition results in \% on the multi-view dataset of OuluVS2 using our proposed method on the given test set per speaker and the corresponding means and standard deviations across speakers (with SC = Sentence correctness, WC = Word correctness and WA = Word accuracy).
}
\centering
\renewcommand{\arraystretch}{1.3}
\begin{tabular}{@{}crrrcrrrcrrrcrrrcrrr@{}}  \toprule
& \multicolumn{3}{c}{$0^\circ$} & \phantom{a} & \multicolumn{3}{c}{$30^\circ$} & \phantom{a} & \multicolumn{3}{c}{$45^\circ$} & \phantom{a} & \multicolumn{3}{c}{$60^\circ$} & \phantom{a} & \multicolumn{3}{c}{$90^\circ$}\\
\cmidrule{2-4} \cmidrule{6-8} \cmidrule{10-12} \cmidrule{14-16} \cmidrule{18-20}
Spkr. & SC & WC & WA && SC & WC & WA && SC & WC & WA && SC & WC & WA && SC & WC & WA\\
\midrule
6 & 73.3 & 73.3 & 73.3 && 70.0 & 70.0 & 73.3 && 53.3 & 60.0 & 56.0 && 40.0 & 40.0 & 37.3 && 50.0 & 57.3 & 52.0\\
8 & 56.7 & 54.7 & 54.7 && 66.7 & 66.7 & 62.7 && 66.7 & 74.7 & 73.3 && 66.7 & 66.7 & 66.7 && 63.3 & 66.7 & 60.0\\
9 & 43.3 & 45.3 & 41.3 && 53.3 & 53.3 & 56.0 && 56.7 & 54.7 & 50.7 && 43.3 & 45.3 & 40.0 && 36.7 & 36.0 & 24.0\\
15 & 73.3 & 77.3 & 76.0 && 63.3 & 63.3 & 58.7 && 53.3 & 54.7 & 53.3 && 40.0 & 40.0 & 34.7 && 40.0 & 41.3 & 30.7\\
26 & 76.7 & 74.7 & 74.7 && 90.0 & 90.0 & 89.3 && 63.3 & 62.7 & 58.7 && 70.0 & 65.3 & 65.3 && 80.0 & 85.3 & 82.7\\
30 & 73.3 & 80.0 & 78.7 && 76.7 & 76.7 & 77.3 && 90.0 & 92.0 & 92.0 && 73.3 & 82.7 & 80.0 && 73.3 & 84.0 & 74.7\\
34 & 96.7 & 97.3 & 97.3 && 86.7 & 86.7 & 90.7 && 80.0 & 85.3 & 85.3 && 80.0 & 80.0 & 78.7 && 63.3 & 61.3 & 56.0\\
43 & 73.3 & 78.7 & 76.0 && 83.3 & 83.3 & 81.3 && 63.3 & 62.7 & 56.0 && 56.7 & 50.7 & 41.3 && 70.0 & 70.7 & 68.0\\
44 & 80.0 & 81.3 & 81.3 && 86.7 & 86.7 & 88.0 && 80.0 & 78.7 & 78.7 && 93.3 & 97.3 & 97.3 && 50.0 & 52.0 & 48.0\\
49 & 86.7 & 88.0 & 86.7 && 80.0 & 80.0 & 80.0 && 93.3 & 93.3 & 93.3 && 83.3 & 86.7 & 86.7 && 63.3 & 69.3 & 68.0\\
51 & 66.7 & 60.0 & 60.0 && 53.3 & 53.3 & 50.7 && 50.0 & 42.7 & 41.3 && 43.3 & 41.3 & 33.3 && 53.3 & 56.0 & 48.0\\
52 & 76.7 & 78.7 & 76.0 && 96.7 & 96.7 & 94.7 && 56.7 & 62.7 & 60.0 && 70.0 & 68.0 & 65.3 && 66.7 & 77.3 & 69.3\\
\midrule
Mean & 73.1 & 74.1 & 73.0 && 75.6 & 76.8 & 75.2 && 67.2 & 68.7 & 66.6 && 63.3 & 63.7 & 60.6 && 59.2 & 63.1 & 56.8\\
SD & 12.9 & 13.8 & 14.2 && 13.6 & 13.4 & 14.3 && 14.3 & 15.3 & 16.6 && 17.6 & 19.2 & 21.5 && 12.7 & 14.9 & 16.7\\
\bottomrule
\end{tabular}
\end{table}

\subsubsection{Single-View Experiments}

As the obtained results show, on average roughly 81\% of the words are correctly recognized during the cross-validation approach for the frontal view. In addition, we have achieved a word recognition correctness of around 74\% on the test set. Also, we can see that 79\% of sentences are correctly recognized during cross-validation while the performance on the test set is 73\%. The small differences between the word correctness and accuracies indicate that there are only few insertion errors. Comparing our obtained results with the baseline cross-validation results on the same dataset reveals that we have improved the performance with a large margin of roughly 5\%.

The average phrase recognition results for the $30^\circ$ view show similar improvements over the baseline provided. Almost 79\% of all sentences are classified correctly for the cross-validation data --- approximately 3\% more than the baseline --- and around 76\% on the test data. Similarly, for the test set 77\% of all words are correct and the accuracy reaches 75\%, while on the cross-validation these values reach 82\% and 81\%. The other views do not show improvements over the baseline.

Looking into the standard deviation indicated in the figures or the individual test results in Table~\ref{tab:exp4_test}, we can see, however, that there is a large margin between the performance of the best speaker and the worst. This hints at a common problem in visual speech recognition where the variability between speakers is very large.

The frame recognition accuracy is shown in Table~\ref{tab:per_frame}. The per frame results on a phoneme and a viseme basis for the training and test sets are displayed here. The observations are two-fold: First, it can be seen that on a frame level the differences between the different view angles does not seem very big, however, the combination of successive frames proves more successful for the frontal views as described above. Secondly, the phoneme and viseme-based classification show a big difference between the 28 phoneme classes and 12 viseme classes (defined according to \cite{Harte_2015}) due to the similarity between various phonemes represented only by the shape of the lips.

\setlength{\tabcolsep}{1.5pt}
\begin{table}
\caption{
\label{tab:per_frame} Frame recognition accuracy results in \% on the multi-view dataset of OuluVS2 using the LSTM output of our proposed on the given train and test sets across all speakers for phonemes and visemes (visemes defined according to~\cite{Harte_2015}).
}
\centering
\renewcommand{\arraystretch}{1.3}
\begin{tabular}{@{}ccrcrcrcrcr@{}}  \toprule
&& $0^\circ$ & \phantom{a} & $30^\circ$ & \phantom{a} & $45^\circ$ & \phantom{a} & $60^\circ$ & \phantom{a} & $90^\circ$\\
\midrule
\multirow{2}{*}{Train} & Phoneme & 19.5 && 20.1 && 18.0 && 17.7 && 15.7\\
 & Viseme & 34.4 && 32.9 && 33.8 && 33.7 && 30.8\\
\multirow{2}{*}{Test} & Phoneme & 17.2 && 17.7 && 17.1 && 17.2 && 16.1\\
 & Viseme & 30.8 && 30.4 && 32.0 && 31.4 && 29.6\\

\bottomrule
\end{tabular}
\end{table}

\subsubsection{Multiple-View Experiments}

\begin{figure}
\begin{center}
\includegraphics[width=0.7\columnwidth]{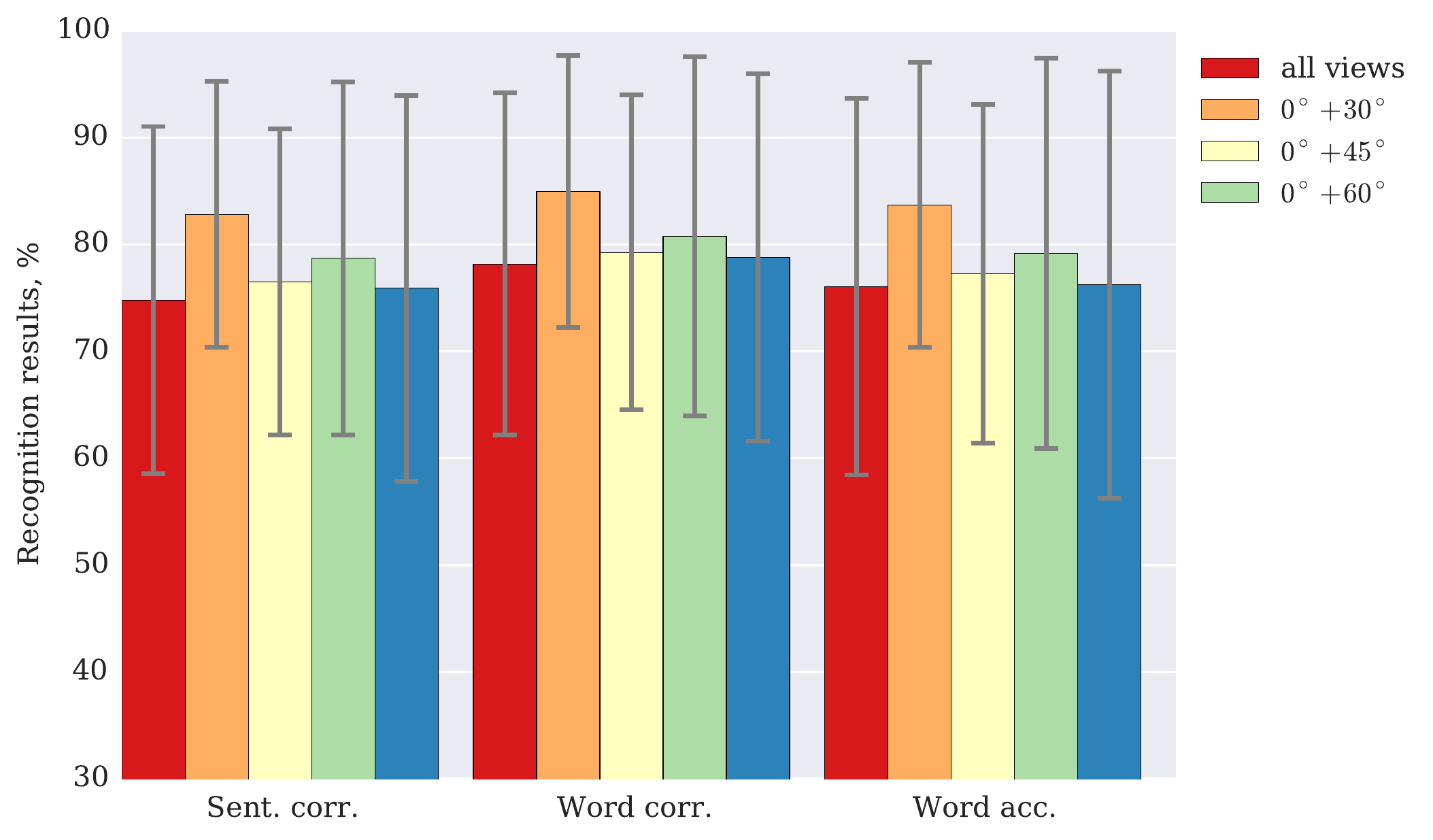}
\caption{{\label{fig:comb_cv}
Mean phrase recognition results on the combination of different views of the multi-view dataset of OuluVS2 using our proposed method with the cross-validation technique and the standard deviation across subjects.%
}}
\end{center}
\end{figure}

\begin{figure}
\begin{center}
\includegraphics[width=0.7\columnwidth]{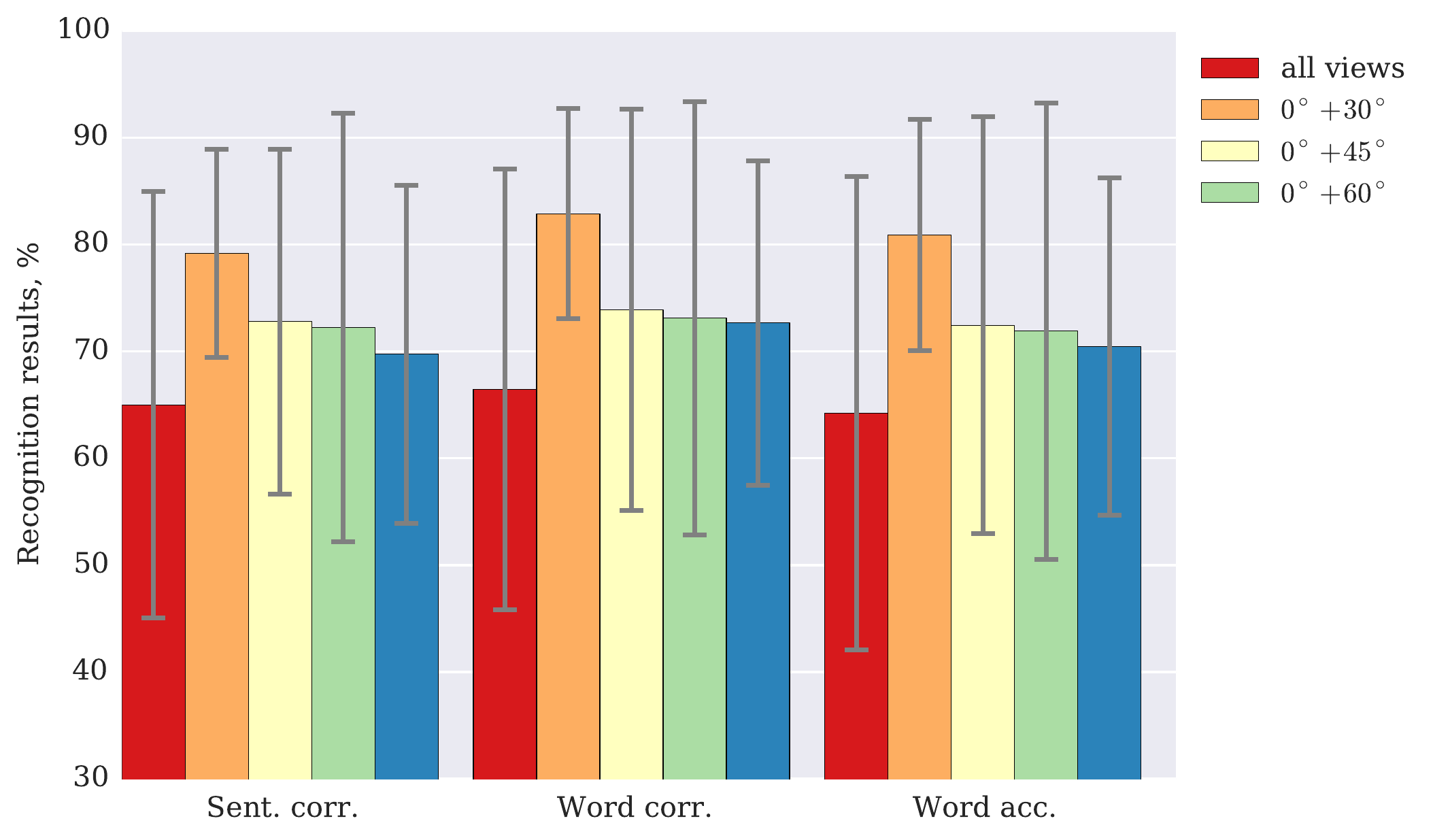}
\caption{{\label{fig:comb_test}
Mean phrase recognition results on the combination of different views of the multi-view dataset of OuluVS2 using our proposed method on the given test set and the standard deviation across subjects.%
}}
\end{center}
\end{figure}

In order to fully benefit from the multi-view recordings, further analyses are performed on combinations of different views. To this end the feature vectors obtained from the LSTM are concatenated and then processed similarly to the single views with their delta and acceleration components in a tandem GMM-HMM system. These multi-view experiments show interesting results (see Figures~\ref{fig:comb_cv} and \ref{fig:comb_test}). Various combinations of the frontal view with each of the four side views were tested, as well as the ensemble of all views together.

The multiple-view results show very good improvements especially for the combination of the frontal and the $30^\circ$-side view. On the cross-validation set a sentence accuracy of nearly 83\% is achieved, while word correctness and word accuracy are around 85\% and 84\% respectively. Thus this amounts to improvements of around 3--10\% over the separate results for these views. Similar improvements can be seen on the test set, where for the same combination the recognition of sentences is at 79\% and 83\% of words are recognised correctly. The word accuracy lies at 81\%. These results show that especially between the frontal and the $30^\circ$-view there is complementary information that can be exploited. The improvements for the other views are not as significant, however, there could be further improvements. The concatenation of all the feature vectors from all views shows a particularly bad result. This is probably due to the increase in dimensionality, which could be aided by prior dimensionality reduction techniques.

Furthermore, again a large variability in the performance between the different speakers can be observed from the standard deviation shown in Figures~\ref{fig:comb_cv} and \ref{fig:comb_test} as well as the individual speaker results in Table~\ref{tab:combined_test1}.

\setlength{\tabcolsep}{0.8pt}
\begin{table}
\caption{
\label{tab:combined_test1} Phrase recognition results in \% on the combination of different views of the multi-view dataset of OuluVS2 using our proposed method on the given test set per speaker and the corresponding means and standard deviations across speakers (with SC = Sentence correctness, WC = Word correctness and WA = Word accuracy).
}
\centering
\renewcommand{\arraystretch}{1.3}
\begin{tabular}{@{}crrrcrrrcrrrcrrrcrrr@{}}  \toprule
& \multicolumn{3}{c}{all views} & \phantom{a} & \multicolumn{3}{c}{$0^\circ + 30^\circ$} & \phantom{a} & \multicolumn{3}{c}{$0^\circ + 45^\circ$} & \phantom{a} & \multicolumn{3}{c}{$0^\circ + 60^\circ$} & \phantom{a} & \multicolumn{3}{c}{$0^\circ + 90^\circ$}\\
\cmidrule{2-4} \cmidrule{6-8} \cmidrule{10-12} \cmidrule{14-16} \cmidrule{18-20}
Spkr. & SC & WC & WA && SC & WC & WA && SC & WC & WA && SC & WC & WA && SC & WC & WA\\
\midrule
6 & 30.0 & 29.3 & 25.3 && 76.7 & 80.0 & 77.3 && 63.3 & 68.0 & 64.0 && 40.0 & 44.0 & 41.3 && 50.0 & 53.3 & 52.0\\
8 & 56.7 & 61.3 & 58.7 && 70.0 & 74.7 & 70.7 && 60.0 & 64.0 & 61.3 && 53.3 & 57.3 & 57.3 && 50.0 & 54.7 & 54.7\\
9 & 36.7 & 44.0 & 38.7 && 70.0 & 76.0 & 73.3 && 43.3 & 44.0 & 41.3 && 60.0 & 64.0 & 62.7 && 43.3 & 54.7 & 45.3\\
15 & 70.0 & 65.3 & 65.3 && 70.0 & 76.0 & 73.3 && 66.7 & 66.7 & 65.3 && 70.0 & 68.0 & 68.0 && 70.0 & 69.3 & 66.7\\
26 & 60.0 & 56.0 & 50.7 && 80.0 & 84.0 & 80.0 && 86.7 & 88.0 & 88.0 && 66.7 & 62.7 & 61.3 && 66.7 & 66.7 & 64.0\\
30 & 76.7 & 82.7 & 81.3 && 86.7 & 90.7 & 90.7 && 83.3 & 88.0 & 88.0 && 90.0 & 93.3 & 93.3 && 83.3 & 90.7 & 88.0\\
34 & 96.7 & 98.7 & 98.7 && 86.7 & 90.7 & 90.7 && 96.7 & 98.7 & 98.7 && 96.7 & 97.3 & 97.3 && 90.0 & 90.7 & 89.3\\
43 & 76.7 & 80.0 & 77.3 && 76.7 & 81.3 & 78.7 && 86.7 & 89.3 & 89.3 && 73.3 & 80.0 & 77.3 && 86.7 & 85.3 & 85.3\\
44 & 70.0 & 72.0 & 72.0 && 83.3 & 88.0 & 88.0 && 76.7 & 81.3 & 78.7 && 100.0 & 100.0 & 100.0 && 70.0 & 70.7 & 70.7\\
49 & 80.0 & 82.7 & 82.7 && 86.7 & 92.0 & 89.3 && 86.7 & 88.0 & 86.7 && 90.0 & 93.3 & 92.0 && 83.3 & 88.0 & 86.7\\
51 & 40.0 & 37.3 & 33.3 && 63.3 & 61.3 & 58.7 && 46.7 & 33.3 & 32.0 && 40.0 & 36.0 & 30.7 && 56.7 & 56.0 & 54.7\\
52 & 86.7 & 88.0 & 86.7 && 100.0 & 100.0 & 100.0 && 76.7 & 77.3 & 76.0 && 86.7 & 81.3 & 81.3 && 86.7 & 92.0 & 88.0\\
\midrule
Mean & 65.0 & 66.4 & 64.2 && 79.2 & 82.9 & 80.9 && 72.8 & 73.9 & 72.4 && 72.2 & 73.1 & 71.9 && 69.7 & 72.7 & 70.4\\
SD & 19.9 & 20.6 & 22.2 && 9.7 & 9.8 & 10.8 && 16.2 & 18.8 & 19.5 && 20.1 & 20.3 & 21.4 && 15.8 & 15.2 & 15.8\\
\bottomrule
\end{tabular}
\end{table}

\section{Conclusion}

In this paper, we have proposed a visual speech recognition system that utilizes a two-stage cascaded PCA network to extract unsupervised learning based lip representations together with a layer of LSTM networks to obtain a set of spatiotemporal visual features. These features have later been used in a tandem GMM-HMM system for speech recognition. As the results indicate, the proposed method has outperformed the baseline technique with a large margin. They also show interesting results in a multiple-view recognition scenario, indicating the complementary information contained in the different views.

In this study only a limited dataset with a small vocabulary has been explored to point out the benefits of using PCA networks in combination with LSTMs. Future works should thus extend this approach to other available datasets such as TCD-TIMIT \cite{Harte_2015} that allow phoneme classification and provide a larger vocabulary. In addition, the influence of the different views and their complementary nature within the framework of these spatiotemporal features could be explored in a more detailed multiple-view visual speech recognition study.

\vspace{3mm}
\noindent {\bf Acknowledgement}. This work was supported by TUBITAK project number 113E067 and by a Marie Curie FP7 Integration Grant within the 7th EU Framework Programme.

\bibliographystyle{splncs03}



\end{document}